\documentclass[runningheads]{llncs}
\usepackage[T1]{fontenc}
\usepackage{graphicx}

\usepackage{diagbox}
\usepackage{booktabs}
\usepackage{cite}
\usepackage[dvipsnames]{xcolor}
\usepackage{wrapfig}
\usepackage{caption}
\usepackage{floatflt}
\usepackage{mathtools}
\usepackage{mathpartir}
\usepackage[linesnumbered, ruled, vlined]{algorithm2e}
\usepackage{caption}
\usepackage{subcaption}
\usepackage{arydshln}
\usepackage{multirow}
\usepackage{slashbox}
\usepackage{tikz}
\usetikzlibrary{automata, positioning}
\usepackage[subtle]{savetrees}

\usepackage{amsmath,amsfonts,amssymb,bm}
\usepackage{stmaryrd}

\def\eqref#1{equation~\ref{#1}}

\def\1{\bm{1}}

\def\vtheta{{\bm{\theta}}}
\def\va{{\bm{a}}}
\def\vb{{\bm{b}}}

\def\vx{{\bm{x}}}

\def\evx{{x}}

\def\mA{{\bm{A}}}

\def\mC{{\bm{C}}}

\def\mM{{\bm{M}}}

\def\mP{{\bm{P}}}

\DeclareMathAlphabet{\mathsfit}{\encodingdefault}{\sfdefault}{m}{sl}
\SetMathAlphabet{\mathsfit}{bold}{\encodingdefault}{\sfdefault}{bx}{n}

 \usepackage{cleveref}
\usepackage{bbm} %

\newcommand\countings[1]{\{1,\ldots,#1\}}
\newcommand\distr[1]{\mathbf{Dist}(#1)} 
\newcommand\subdistr[1]{\mathbf{SDist}(#1)} %
\def\indicator{\mathbbm{1}} %
\newcommand\vIndicator[1]{\va_{#1}} 

\newcommand\norm[1]{\mathit{norm}(#1)} 
\def\dtmc{M}

\def\barsp{\, | \,}
\newcommand\sem[1]{\llbracket #1 \rrbracket}

\def\condition{e}
\def\conditionSet{E}

\def\actual{s}

\def\actualSet{S}

\def\actualSys{\underline{\actualSet}}
\def\actualErr{\actual_{err}}

\def\img{r}

\def\imgSet{R}
\def\imgDataset{\mathtt{R}}
\def\est{y}

\def\estSet{Y}
\def\control{u}

\def\controlSet{U}

\def\fpos{f^\mathit{pos}}

\def\distS{\mathit{pos}_{\mathit{{side}}}}
\def\distF{\mathit{pos}_{\mathit{front}}}
\def\head{\mathit{heading}}

\def\maxt{30}

\def\scenario{w}

\newcommand*{\linfer}[3][]{%
  \inferrule*[right=#1,
              rightskip=2em,
              rightstyle=\sc]%
  {#2}{#3}
}

\usepackage{mdframed}
\newmdenv[
  linecolor=black,
  backgroundcolor=gray!10,
  skipabove=10pt,
  skipbelow=10pt,
  innertopmargin=5pt,
  innerbottommargin=5pt,
  innerleftmargin=10pt,
  innerrightmargin=10pt,
  roundcorner=5pt
]{queryblock}
\begin{document}
\title{Scenario-based Compositional Verification of Autonomous Systems with Neural Perception}

\titlerunning{Scenario-based Compositional Verification of Autonomous Systems}
\author{Christopher Watson\inst{1} \and
Rajeev Alur\inst{1} \and
Divya Gopinath\inst{2}
Ravi Mangal\inst{3} \and \\
Corina S. P\u{a}s\u{a}reanu\inst{2,4}}
\institute{University of Pennsylvania, USA \and
KBR Inc., NASA Ames, USA \and
Colorado State University, USA \and
Carnegie Melon University, USA}
\authorrunning{C. Watson et al.}
\maketitle              %

\begin{abstract}
Recent advances in deep learning have enabled the development of autonomous systems that use deep neural networks for perception. 
Formal verification of these systems is challenging due to the size and complexity of the perception DNNs as well as hard-to-quantify, changing environment conditions.
To address these challenges, we propose a probabilistic verification framework for autonomous systems based on the following key concepts:
(1) {\em Scenario-based Modeling:} We decompose the task (e.g., car navigation) into a composition of {\em scenarios}, each representing a different environment condition.
(2) {\em Probabilistic Abstractions:} For each scenario, we build a compact {\em abstraction of perception} based on the DNN's performance on an offline dataset that represents the scenario's environment condition.
(3) {\em  Symbolic Reasoning and Acceleration:} The abstractions enable efficient compositional verification of the autonomous system via symbolic reasoning and a novel {\em acceleration proof rule} that bounds the error probability of the system under arbitrary variations of environment conditions.
We illustrate our approach on two case studies: an experimental autonomous system that guides airplanes on taxiways using high-dimensional perception DNNs and a simulation model of an F1Tenth autonomous car using LiDAR observations. 
\end{abstract}

\section{Introduction}\label{sec:intro}

Recent advances in deep learning have enabled the development of autonomous systems that use deep neural networks (DNNs) for perception~\cite{fremont-cav20,KadronGPY21, abs-1910-07738,abs-1904-00649,huang2020survey,fremont-itsc20}
Formal verification of these systems is uniquely challenging
due to the complexity of formal reasoning about the DNNs. 
The perception DNNs are massive (with millions or billions of parameters) and are also known to be very sensitive to input perturbations. For instance, changing the light conditions in an image can lead to very different DNN outputs and this can adversely affect the safety of the overall system.
Furthermore, real-world autonomous systems typically operate in an environment whose configuration is a priori unknown (e.g., the car is navigating in a location with an unknown layout of tracks).
In this work, we propose a probabilistic verification framework to address these challenges.

\subsubsection{\bf System Description.} 
\begin{wrapfigure}{r}{0.5\textwidth}
\centering
 \vspace{-0.6cm}
\includegraphics[width=\linewidth]{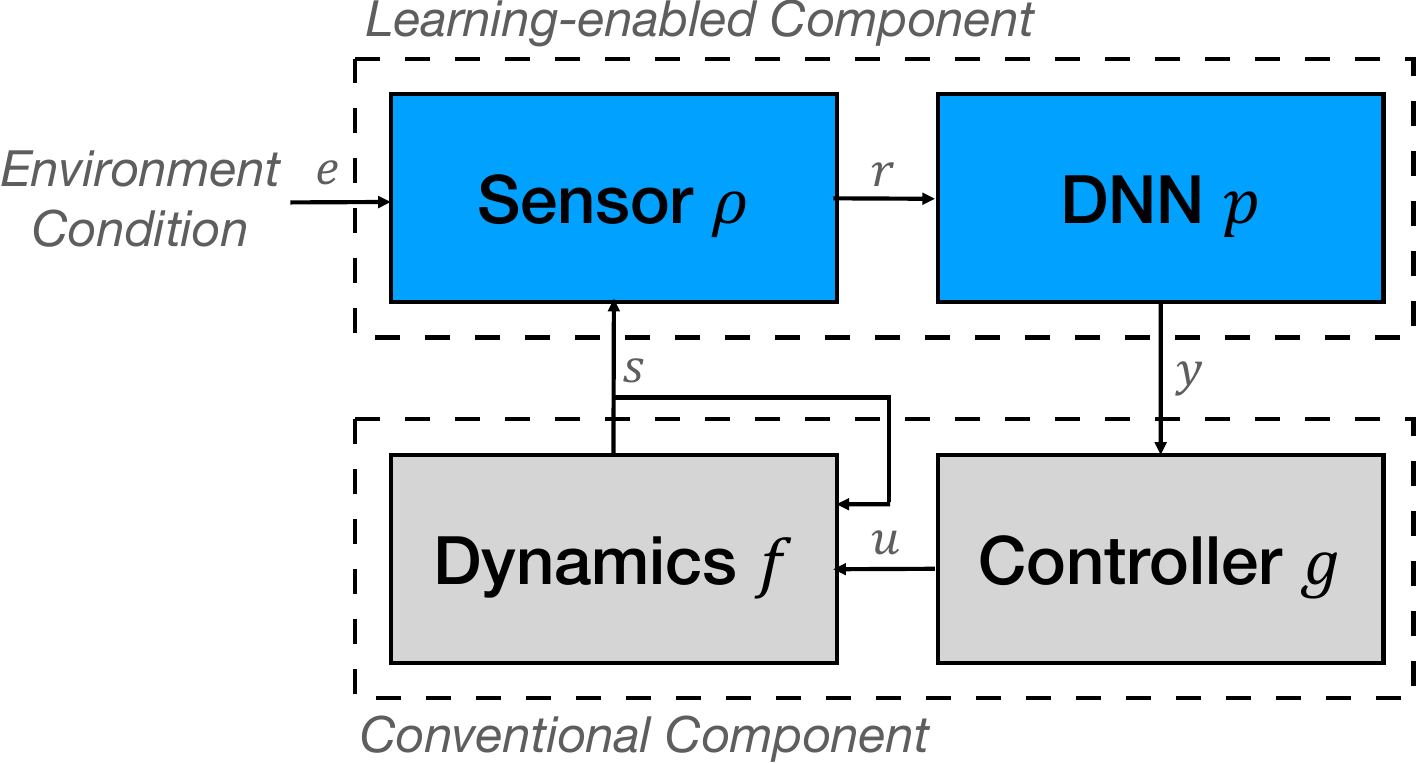}
\caption{\small Components of an autonomous system with DNN-based perception.}\vspace{-0.6cm}
\label{fig:components}
\end{wrapfigure}

Consider the autonomous system from Figure~\ref{fig:components}; it consists of four interacting components: \textit{Sensor}, \textit{Perception DNN}, \textit{Controller}, and \textit{Dynamics}. The \textit{Sensor} (e.g., a camera or LiDAR system) observes the current underlying system state (potentially subject to hard-to-model environmental conditions) and produces sensor readings (e.g., images or LiDAR readings).
 The \textit{Perception DNN} takes inputs from the \textit{Sensor} and produces {\em state estimates} that are used by the \textit{Controller} to output control commands for the system. In response to these commands, the system state is updated as described by the \textit{Dynamics} and the cycle repeats. This forms a closed-loop system where the \textit{Perception DNN} repeatedly receives \textit{Sensor} inputs as the system operates in its environment.
Note that such systems tend to be stochastic, even when \textit{Dynamics} and \textit{Controller} are deterministic, due to the uncertainty in the \textit{Sensor} readings.

\noindent{\bf Safety Properties.} We analyze such systems with respect to critical safety properties that can be framed as reachability queries.
We focus on two questions: (1) 
given a set of initial conditions, what is the probability of reaching an error state? and (2) given a user-specified bound on the probability of reaching an error state, what is the corresponding set of most permissive initial conditions?
The specific notion of error depends on the system being analyzed. In our first case study, we analyze an autonomous airplane taxiing system that relies on Boeing's \textit{TaxiNet} perception DNN and define error as excessive deviation from the runway's centerline. In our simulated F1Tenth case study, we define error to be collision with the edge of the race track or failure to navigate a track segment within a predetermined time limit.

\noindent{\bf Modeling the Learning-enabled Components.}
The conventional components (\textit{Controller} and  \textit{Dynamics}) can be modeled  using well-known techniques~\cite{pasareanu2023assumption}, for instance, they can be modeled as transition systems or  Discrete-time Markov Chains (DTMCs).
However, the learning-enabled components (\textit{Perception DNN} and \textit{Sensor}) are difficult to model due to enormous sizes of modern DNNs and the complex nature of the sensors as well as the changing environmental conditions in which these systems operate.
The key challenge is to analyze the learning enabled components in a way that can be composed with traditional systems verification techniques to obtain a guarantee for the entire closed loop system.
Approaches to this challenge form three groups: 
(1) those that apply neural network verification to the \textit{Perception DNN} and precisely model the \textit{Sensor} mathematically~\cite{Pavithra23,10384107,10.1007/978-3-031-06773-0_11,Pavithra24} or with a learned model~\cite{katz2022verification,cai2024scalable,arjomandbigdeli2024saiv,waite2023datadriven} 
(2) those that build \textit{contracts}~\cite{EMSOFT22-Hsieh,perceptioncontracts,pasareanu2023assumption,li2023refining,landing-ICRA24,formationACC24, Li-et-al} 
around the learning enabled components, and
(3) those that build \textit{probabilistic abstractions}~\cite{badithela2021leveraging, pasareanu2023closed, badithela2023evaluation, calinescu2022discrete, calinescu2024controller} around the learning enabled component.

In general, neural network verification only scales to DNNs of modest size, and realistic sensors are hard to model, especially when subject to environmental disturbances.
Contract-based approaches provide obligations for the learning-enabled components that ensure system safety, but cannot quantify system safety if the contracts are not met.
To provide a scalable, quantitative analysis of sytem-level safety, our current work extends the technique of \textit{probabilistic abstraction}, which models the learning-enabled components' behavior as an explicit map from each system state to a distribution over \textit{Perception DNN} outputs.
The probabilities in this map are computed from the {\em confusion matrix}  that measures the performance of the \textit{Perception DNN} on a representative dataset obtained from simulations or real-world setting.
With this abstraction, the resulting system  can be modeled as a DTMC and is amenable to verification using off-the-shelf tools such as PRISM~\cite{kwiatkowska2011prism} and Storm~\cite{STORM}. 

While the compact abstraction of the learning-enabled components allows the analysis to scale to arbitrarily large DNNs and does not require modeling the \textit{Sensor}, if the offline dataset pools data from multiple distinct environment conditions the abstraction may be too coarse, resulting in an imprecise analysis.%

\begin{wrapfigure}{r}{0.5\textwidth}
    \centering
    \vspace{-1cm}
    \includegraphics[width=1\linewidth]{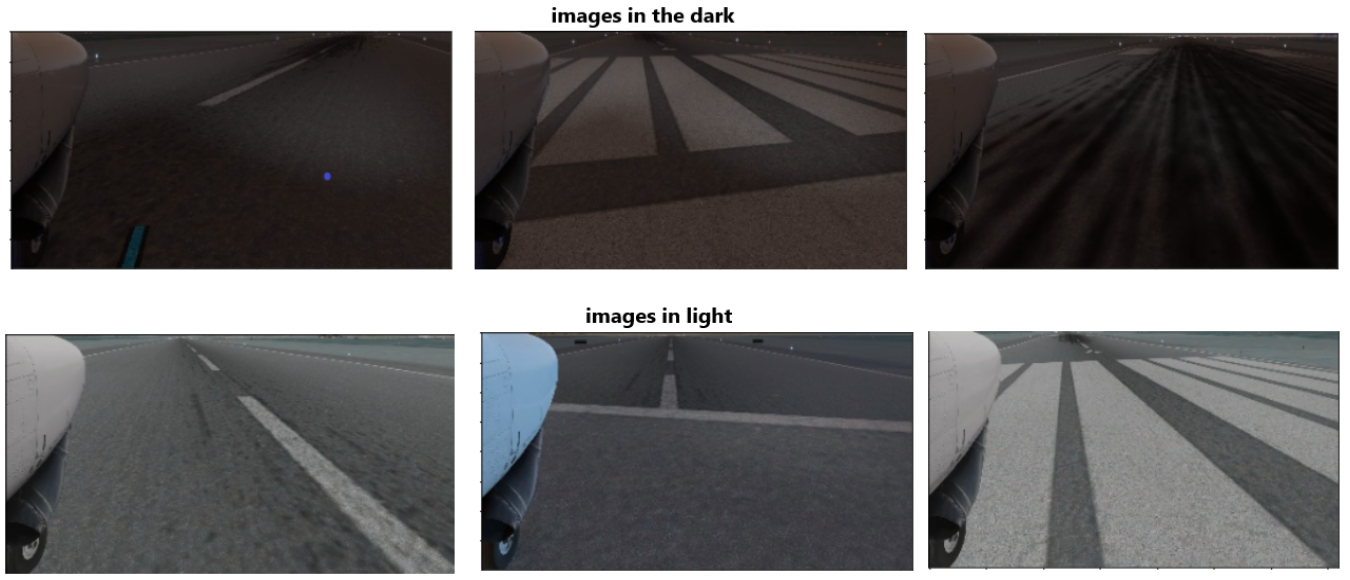}
    \vspace{-0.5cm}
    \caption{\small
    TaxiNet achieves 91.25\% accuracy when estimating heading for a \textit{bright} image dataset but only 53.87\% accuracy for a \textit{dark} image dataset.
    }\vspace{-0.5cm}
    \label{fig:night_day_examples}
\end{wrapfigure}

\noindent{\bf Our Proposal.}
In this work, we structure the analysis of the autonomous system as a composition of \textit{scenarios}, each of which represents the behavior of the system under a different \textit{environment condition} for a fixed duration.
In the context of our autonomous taxiing case study, we observe that the \textit{TaxiNet} perception system has dramatically different performance in \textit{bright} versus \textit{dark} lighting conditions (see \Cref{fig:night_day_examples}). 
To perform an analysis of system level safety that accounts for changes in lighting conditions, we define the scenarios \texttt{bright} and \texttt{dark}.
Within the \texttt{bright} scenario, we can accurately model the behavior of the closed loop system by incorporating a probabilistic of abstraction constructed using a dataset collected in \textit{bright} light. 
A separate dataset lets us accurately model the system's behavior in the \texttt{dark} scenario.
Similarly, to analyze an autonomous race car, we define a separate scenario for each kind of track segment (\textit{straight}, \textit{left turn}, or \textit{right turn}) to efficiently analyze safety for a diverse set of track configurations. While scenario-based decomposition is inspired by the non-probabilistic analysis of~\cite{ivanov2021compositional}, our probabilistic abstraction of the learning-enabled component of each scenario allows our analysis to scale to arbitrarily complex sensor and perception systems. Scenario-based decomposition also allows our approach to efficiently compute safety guarantees for long time horizons.

\subsubsection{Contributions.} Overall, we make the following contributions:
(1) we present an approach for compactly modeling autonomous systems that accounts for changing environment conditions via decomposition into environment-specific scenarios;
(2) we extend prior work \cite{pasareanu2023closed} by allowing a system designer to define separate probabilistic abstractions, representing DNN performance in different environment conditions; we further generalize the confusion matrices to account for arbitrary state estimate definitions, not just the underlying system state as in previous work;
(3) 
we describe an efficient symbolic approach for analyzing finite sequences of scenarios with respect to initial conditions specified as a set of distributions over the system states; (4) we also give a novel family of compositional proof rules for accelerated analysis of finite sequences with arbitrary interleavings of scenarios; these rules enable developers to compute a bound on error probability even when the precise configuration of the environment is not known a priori; (5) we illustrate our approach on two
case studies: a model of an experimental autonomous system that guides airplanes on taxiways using high-dimensional perception DNNs and a model of an F1Tenth autonomous car using LiDAR observations.

\subsubsection{Related Work.}
We have already discussed closely related work on formal analysis of learning-enabled systems. Other related work includes approaches based on statistical simulation \cite{Yalcinkaya-et-al} that require less computation but provide a different type of guarantee than probabilistic model checking. Moreover, statistical model checking is not suitable for models that include nondeterminism, thus cannot be compared with the guarantees we obtain for arbitrary interleavings of scenarios. Another related line of work is compositional probabilistic model checking \cite{Watanabe-et-al}, which, however does not address the central challenge of incorporating machine-learning components.

Also related is the work on probabilistic predicate transformers \cite{PredTransf}.  Each of our scenarios can be seen as a probabilistic assignment, and the backward analysis of Section \ref{sec:analysis-sequential} is a special case of the weakest precondition computation. Much work on predicate transformers concerns general loops, which are not immediately relevant to modeling autonomous systems over bounded horizons. To our knowledge, no analogue of our acceleration rules exists in the literature.

\section{Scenario-based Probabilistic Abstractions of Perception}%
\label{sec:approach}

\begin{figure}[t]
    \begin{center}

    \includegraphics[width=\textwidth]{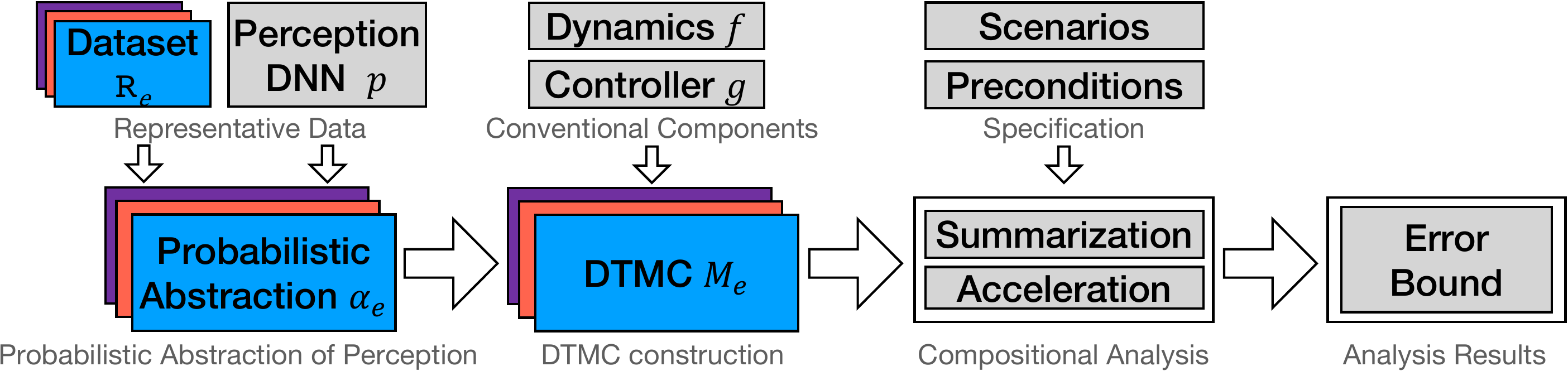}
    \end{center}\vspace{-0.5cm}
    \caption{\small Scenario-based compositional analysis of autonomous systems.}\vspace{-0.5cm}
    \label{fig:framework}
\end{figure}

Figure~\ref{fig:framework} illustrates our approach. The goal is to analyze safety properties of autonomous systems, subject to changing environment conditions.
We assume that a domain expert can identify a finite set $\conditionSet$ of environment conditions, e.g. \textit{bright} vs \textit{dark} lighting. For each environment condition $\condition \in \conditionSet$, the domain expert provides a dataset $\imgDataset_\condition$ that represents the behavior of the \textit{Sensor} subject to that environment condition. This, combined with black-box access to the \textit{Perception DNN}, allows us to construct a \textit{probabilistic abstraction of perception} $\alpha_\condition$ for each environment condition, which can in turn be composed with a model of the conventional components to yield a DTMC model $M_{\condition}$ of the closed-loop system subject to each environment condition. For simplicity of presentation, we assume that only the behavior of learning-enabled components changes with the environment conditions, more generally, the conventional components may also change. Our techniques are applicable to this more general setting, and in Section~\ref{sec:experiments} we consider such a case.

A scenario represents the behavior of the autonomous system subject to a particular environment condition for a fixed finite time horizon. 
The inputs to our compositional analysis are a sequence of scenarios and a precondition that constrains the set of initial state distributions. A \textit{summarization} procedure allows us to efficiently compute a worst-case error bound for a fixed scenario sequence, and our \textit{acceleration} proof rules allow us to derive a bound on error probability that is parametric with respect to the length of a scenario sequence of unbounded length.
We make these notions precise in the following sections.

\subsection{Modeling the Closed-loop Autonomous System}

A discrete-time closed-loop autonomous system is composed of conventional components \textit{Controller} and \textit{Dynamics} as well learning-enabled components \textit{Sensor} and \textit{Perception DNN}. Given a fixed environment condition $e\in E$, at each timestep $t$, the \textit{Sensor} $\rho_\condition: \actualSet \rightarrow \distr{\imgSet}$\footnote{We write $\distr{X}$ for the set of probability distributions over a finite set $X$ and identify each $\vx \in \distr{X}$ with its representation as a row vector in $\mathbb{R}^{|X|}$.} (which may be a camera or LiDAR system) observes the current underlying system state $\actual_t \in \actualSet$ and probabilistically produces a sensor reading $\img_t  \in \imgSet$ sampled from the distribution $\rho_\condition(\actual_t)$ (where $\img_t$ is an image, LiDAR reading, etc).
The \textit{Perception DNN} $p: \imgSet \rightarrow \estSet$ processes $\img_t$ to yield an estimate $\est_t \in \estSet$ of the system state. 
The \textit{Controller} $g:\estSet \rightarrow \controlSet$ receives this state estimate and outputs a control signal $\control_t \in \controlSet$.  At the start of the next timestep $t+1$, the \textit{Dynamics} $f:\actualSet \times \controlSet \rightarrow \actualSet$ produces a new system state $\actual_{t+1} \in \actualSet$ in response to the control input $\control_t$ and the cycle continues. The sets $\actualSet$, $\estSet$, and $\controlSet$ are assumed to be finite. 

\subsubsection{Probabilistic Abstraction of Perception.} 
Constructing models of the \textit{Sensor} and the \textit{Perception DNN} is a central challenge given their complexity. We extend past work~\cite{pasareanu2023closed} and  build  compact abstractions $\alpha_e:\actualSet \rightarrow \distr{\estSet}$ of the learning-enabled components.

The key idea is that the input-output behavior of the learning-enabled components (as a function of type $\actualSet \rightarrow \distr{\estSet}$) can be estimated using  an offline dataset of sensor readings gathered using the \textit{Sensor} onboard the autonomous system. 
A limitation of previous work was that they built a single abstraction, regardless of environment conditions, leading to overly imprecise results, as illustrated in Figure~\ref{fig:night_day_examples}.

In this work, we address this important limitation by 
building a separate probabilistic abstraction of perception $\alpha_e$ for each environment condition $e$. For each pair of environment condition $e$ and state $\actual$, we assume access to an offline dataset $\imgDataset_{\condition, \actual}$ of sensor readings collected while the autonomous system was in state $\actual$ and subject to condition $\condition$. 

Another limitation of previous work was that the probabilistic abstraction was based on computing confusion matrices under the restricted assumption that the DNN outputs correspond precisely to the state definition. In practice, DNNs can be used to estimate only some aspects of the underlying state. Thus, we generalize confusion matrices to account
for arbitrary state estimate definitions, not just the underlying system state as
in previous work.

The construction of probabilistic abstraction $\alpha_{\condition}$ for condition $\condition$ proceeds as follows.
First, for each $\actual \in \actualSet$, we apply the \textit{Perception DNN} function $p : \imgSet \rightarrow \estSet$ to each $\img_s$ in 
$\imgDataset_{\condition,\actual}$ to obtain the pair $(\actual, p(\img_s))$; here, $\img_s$ is a sensor reading gathered in state $s$ because it is from  $\imgDataset_{\condition,\actual}$. We combine all of these pairs to form a labeled dataset of $\actualSet \times \estSet$ pairs that describes the behavior of the learning-enabled components subject to $\condition$. 
From this dataset we can build a contingency matrix\footnote{This is a \textit{confusion matrix} when $\actualSet = \estSet$.} $\mC_\condition$. 
For any $\actual \in \actualSet$ and $\est \in \estSet$, we write $\mC_\condition(\actual, \est)$ for the frequency of the pair $(\actual, \est)$ in the labeled dataset. We define the probabilistic abstraction of perception $\alpha_\condition : \actualSet \rightarrow \distr{\estSet}$ as 
\begin{equation*}
    \alpha_\condition(\actual)(\est) \coloneqq \frac{\mC_\condition(\actual, \est)}{\sum_{\est' \in \estSet}\left(\mC_\condition(\actual, \est') \right)}
\end{equation*}

\subsubsection{System Model.}

The probabilistic abstraction $\alpha_\condition$ of the learning-enabled component in environment condition $\condition$ can be composed with the discrete models of the conventional components to yield a discrete-time Markov chain (DTMC) that models the closed-loop system as in~\cite{pasareanu2023closed,calinescu2024controller, badithela2021leveraging}.
We denote this DTMC as $\mathcal{M}_\condition = (\actualSet, \mP_\condition)$ in which $\actualSet$ is a finite set of states (in particular, the set of states of the autonomous system) and $\mP_\condition$ is a $|\actualSet| \times |\actualSet|$ right-stochastic matrix. At each timestep, the probability of a transition from some $\actual \in \actualSet$ to some $\actual' \in \actualSet$ is:
$$\mP_\condition(\actual,\actual') \coloneqq \sum_{\est \in \estSet}\alpha_\condition(\actual)(\est)[f(g(\est)) = \actual']$$
where $[f(g(\est)) = \actual']$ evaluates to $1$ whenever $f(g(\est)) = \actual'$ and $0$ otherwise. 
For our case studies, we follow the procedure of~\cite{pasareanu2023closed} to represent this DTMC using PRISM, see~\cref{sec:implementation} for details. 

Without loss of generality, we assume that the autonomous system has a distinguished \textit{error state}, which we denote $\actualErr$. 
We assume that the conventional dynamics ensure that the error state is \textit{absorbing}, i.e., once the system reaches $\actualErr$, it stays there. This modeling assumption both reflects the natural meaning of an ``error state'' and facilitates the compositional analysis.%

\subsection{Scenarios}\label{sec:scenario}
We are interested in modeling the behavior of autonomous systems under changing environmental conditions. For this purpose, we introduce the notion of \textit{scenarios}.
Consider an autonomous system with states $\actualSet$ and environment conditions $\conditionSet$. For each $\condition \in \conditionSet$, let $\dtmc_\condition = (\actualSet, \mP_\condition)$ be the DTMC that models execution of the autonomous system subject to environment condition $\condition$.

A \textit{scenario} is a pair $(\condition, H) \in \conditionSet \times \mathbb{N}$ that comprises an environment condition and a time horizon. Semantically, each scenario induces a map $\sem{(\condition, H)}: \distr{\actualSet} \rightarrow \distr{\actualSet}$ defined as 
$$\sem{(\condition, H)}(\vx) \coloneqq \vx\mP_\condition^H $$
that maps a distribution $\vx \in \distr{\actualSet}$ to the transient distribution reached after taking $H$ steps in $\dtmc_\condition$.

To model transitions between environment conditions, we consider non-empty sequences of scenarios. Each scenario is also a \textit{scenario sequence}. Given two scenario sequences $\scenario_1$ and $\scenario_2$, the sequential composition $\scenario_1;\scenario_2$ is also a scenario sequence. Sequential composition is associative. Semantically, we define $\sem{\scenario_1 ; \scenario_2} : \distr{\actualSet} \rightarrow \distr{\actualSet}$ inductively as:
$$\sem{\scenario ; \scenario'} \coloneqq \sem{\scenario'} \circ \sem{\scenario}$$

\paragraph{\textbf{Scenario Summary.}} When representing the semantics $\sem{\scenario}$ of a scenario $\scenario = (\condition, H)$,
we find it useful to treat the error state $\actualErr$ in a special manner as this facilitates the compositional analysis presented in \Cref{sec:analysis-compositional}.
To treat the error state separately, we first partition the state space $\actualSet$ of $M_{\condition}$ as $\actualSet = \actualSys \sqcup \{\actualErr\}$. Then, our explicit representation of $\sem{(\condition, H)}$ takes the form of the \textit{summary} $(\mA, \vb)$, such that for any initial distribution $\vx \in \distr{\actualSys}$
(we assume here that the initial distribution places no weight on $\actualErr$), executing the scenario $(\condition, H)$ causes the autonomous system to transition to the error state with probability $\vx \cdot \vb$ and induces the subdistribution\footnote{A subdistribution $\vx \in \subdistr{\actualSys}$ may have $|\vx| < 1$.} $\vx\mA \in \subdistr{\actualSys}$ over the non-error states.
Since we assume that the error state is absorbing, the error probability $\vx \cdot \vb$ is the probability that the system reaches the error state at any timestep during the scenario. The summary $(\mA, \vb)$ of an atomic scenario $\scenario = (\condition,H)$ is related to the stochastic matrix $\mP_{\condition}$ as follows:
\[ 
\begin{bmatrix}
\mA & \vb \\
\mathbf{0} & 1 \\
\end{bmatrix}
= \mP_{\condition}^H
\]

We lift the definition of summary to scenario sequences inductively. If a scenario $w$ has summary $C$, then the singleton scenario sequence $w$ also has summary $C$. 
If the scenario sequences $w_1$ and $w_2$ have summaries $C_1 = (\mA_1, \vb_1)$ and $C_2 = (\mA_2, \vb_2)$, respectively, then the scenario sequence $\scenario_1;\scenario_2$ has the summary $C_1C_2 = (\mA_1\mA_2, \vb_1+\mA_1\vb_2)$. This composition is well-defined because each summary is defined with respect to the set of system states $\actualSet$, which is common to all scenarios.

\section{Compositional Analysis  with Scenario Summaries}\label{sec:analysis}

We analyze the probability that the autonomous system will reach the distinguished error state $\actualErr$ during the execution of a scenario sequence. 
We propose two types of analysis: 
(1) a symbolic analysis that provides a tight analysis of the error probability of a fixed scenario sequence with respect to a symbolic precondition and (2) a novel \textit{acceleration rule} that can be used to derive an upper bound on error probability that is parametric with respect to the length of an arbitrary interleaving of a set of scenarios.

\subsection{Symbolic Analysis of Fixed Scenarios}\label{sec:analysis-sequential}

For a fixed scenario sequence $w$ with summary $(\mA, \vb)$, the expression $\vx \cdot \vb$ represents the probability that the autonomous system will reach the error state when started from an initial state distribution $\vx$. This linear expression allows us to efficiently perform \textit{forward} and \textit{backward} analyses that relate preconditions over initial distributions with bounds on error probability.

\subsubsection{Forward analysis.} 
A domain expert may identify a precondition $\phi$ over the set of initial state distributions and wish to compute the worst-case probability of reaching the error state during the execution of the scenario sequence from an initial distribution $\vx \in \distr{\actualSys}$ that satisfies $\phi$.
Recall that we represent each distribution as a vector in $\mathbb{R}^{|\actualSys|}$. We consider preconditions that can be expressed as an intersection of affine constraints, i.e. of the form $\vx\va_1 \leq \theta_1 \land \ldots \land \vx\va_d \leq \theta_d$ for some coefficient vectors $\va_1, \ldots, \va_d \in \mathbb{R}^{|\actualSys|}$ and offset scalars $\theta_1, \ldots, \theta_d \in \mathbb{R}$.

The worst-case error probability is given by the following linear program:
\begin{equation*}
    \begin{aligned}
        \max_{\vx} \quad & \vx \cdot \vb \\
        \text{subject to} \quad &\vx \cdot \va_i \leq \theta_i, \quad \forall i, \in \countings{d}\\
                                & \vx \geq 0,\\ %
                                & \|\vx\|_1 = 1.
    \end{aligned}
\end{equation*}
where the first $d$ constraints encode $\phi$ and the last two constraints enforce that $\vx$ represents a probability distribution. This linear program can be solved using optimization software such as Gurobi~\cite{gurobi}. %

\subsubsection{Backward analysis.}
If a domain expert provides a maximum allowable error probability $\epsilon$ for the execution of the scenario sequence, the precondition
\hbox{$\phi \coloneqq \vx \cdot \vb \leq \epsilon$}
is the weakest precondition over initial state distributions that ensures the error probability does not exceed $\epsilon$.

\subsubsection{Comparison with PRISM.} The symbolic analyses allow the domain expert to reason about the error probability of the autonomous system with respect to a set of initial distributions. This complements the analyses supported by the probabilistic model checker PRISM, which permits forward analysis from a single initial state (or finite set of initial states).

\subsection{Acceleration Rules}\label{sec:analysis-compositional}

The aforementioned analyses are useful when the domain expert fixes a particular scenario sequence to consider. However, such foresight is not always possible. To address this, we introduce novel \textit{acceleration} rules that allow us to bound the probability of error without fixing the length of the scenario sequence or the order in which the scenarios appear \textit{a priori}. 
with respect to an unpredictable sequence of environment conditions. 

\subsubsection{Hoare-style assertions.}
We reason about scenario sequences using Hoare-style assertions with special treatment of error states. An assertion is a quadruple of the form $\{\phi\}C\{\psi\}\{\epsilon\}$
where $C \equiv (\mA,\vb)$ denotes the summary of a scenario sequence,
the precondition $\phi$ and postcondition $\psi$ are predicates over $\distr{\actualSys}$ expressed as intersections of affine constraints over $\mathbb{R}^{|\actualSys|}$ and $0 \le \epsilon \le 1$ is the required bound on error probability. 
We write $\phi(\vx)$ when $\vx$ satisfies $\phi$.
The assertion $\{\phi\}C\{\psi\}\{\epsilon\}$ holds exactly when
$$\forall \vx \in \distr{\actualSys}, \phi(\vx) \Rightarrow \vx \cdot \vb \leq \epsilon \land \psi(\mathit{norm}(\vx\mA))$$
Here $\mathit{norm}$ is an operation that (L1) {\em normalizes} a subdistribution to make it a proper distribution, i.e.,
$norm(\vx)=\frac{\vx}{|\vx|}$
We normalize the output sub-distribution $\vx\mA$ of $C$ as $\psi$ is defined with respect to $\distr{\actualSys}$. Given the explicit summary of the scenario, such assertions can be checked efficiently using an off-the shelf solver, such as Z3~\cite{Z3DeMoura} as we will show in the following example.

\subsubsection{Example.} Consider the scenario with summary $C \equiv (\mA,\vb)$ shown below:

\begin{minipage}{0.45\textwidth}
\[
\mA = 
\begin{bmatrix}
0.6 & 0.2 \\
0.2 & 0.7
\end{bmatrix}\;
\vb = 
\begin{bmatrix}
0.2 \\
0.1 
\end{bmatrix}
\]
\end{minipage}%
\hfill
\begin{minipage}{0.5\textwidth}
\centering
\begin{tikzpicture}[->, >=stealth, auto, node distance=1.9cm, semithick]
  \node[state] (s1) {$s_1$};
  \node[state] (s2) [right of=s1] {$s_2$};
  \node[state] (err) [below of=s2] {$\actualErr$};
  \path (s1) edge [loop left] node {0.6} (s1)
        (s1) edge [bend left] node {0.2} (s2)
        (s1) edge [bend right] node [left] {0.2} (err)
        (s2) edge [loop right] node {0.7} (s2)
        (s2) edge [bend left] node {0.2} (s1)
        (s2) edge node {0.1} (err)
        (err) edge [loop right] node {1} (err);
\end{tikzpicture}
\end{minipage}
Let $\vx = [\evx_1,\evx_2]$ denote an initial distribution over the system states $\actual_1$ and $\actual_2$. Upon executing the scenario, the probability of transitioning to error state is $\vx \cdot \vb = 0.2\evx_1 + 0.1\evx_2$ and the resulting distribution over the system states (conditioned on not reaching $\actualErr$) will be:
$$ \vx' = \mathit{norm}(\vx\mA) = \frac{\vx\mA}{1-\vx\cdot\vb}= \frac{[0.6\evx_1+0.2\evx_2,0.2\evx_1+0.7\evx_2]}{(1-(0.2\evx_1 + 0.1\evx_2))}$$
Let us consider the precondition
$\phi \coloneqq \evx_1 \le 0.7 \land \evx_2 \le 0.7$.
We will show how to check the assertion $\{\phi\}C\{\phi\}\{ 0.15\}$ using Z3.

We can encode the set of distributions that satisfy $\phi$ as a polyhedron in $\mathbb{R}^{|\actualSys|}$ with H-representation $\mM_\phi\vx^\top \le \theta_\phi$ for some $6 \times 2$ matrix $\mM_\phi$ and $6$-dimensional vector $\vtheta_\phi$ of offsets. The H-representation comprises $6$ inequalities since four inequalities define the subset of $\mathbb{R}^{|\actualSys|}$ that represents $\distr{\actualSys}$ and the predicate $\phi$ contains two additional constraints. 
In order to check whether $\phi(\vx) \Rightarrow \vx \cdot \vb \leq \epsilon \land \psi(\mathit{norm}(\vx\mA))$ we can ask Z3 whether 
$\mM_\phi\vx^\top \le \vtheta_\phi \land \neg( \vx \cdot \vb \leq 0.15)$
and 
$\mM_\phi\vx^\top \le \vtheta_\phi \land \neg (\mM(\mathit{norm}(\vx\mA))^\top \leq \vtheta_X)$
are both unsatisfiable. If both are unsatisfiable (as they are for this example), then we can conclude $\{\phi\}C\{\phi\}\{ 0.15\}$.

\subsubsection{Rule for sequential composition.} 
The building block for our acceleration rules is the following compositional rule which bounds the error probability for the sequential composition of two scenarios.

$$
  \linfer[Rule 1]{\{\phi\}C\{\psi\}\{\epsilon\}, \{\phi'\}C'\{\psi'\}\{\epsilon'\}, \psi\implies \phi'}
  {\{\phi\}C C'\{\psi'\}\{1-(1-\epsilon)(1-\epsilon')\}} 
$$

Rule 1 generalizes to our first {\em acceleration rule}, where $C^k$ is shorthand for sequential composition of $C$ applied $k \in \mathbb{N}$ times

$$
  \linfer[Rule 2]{\{\phi\}C\{\phi\}\{\epsilon\}}
  {\{\phi\}C^k\{\phi\}\{1-(1-\epsilon)^k\}} 
$$

Rule 2 can be seen as a family of rules (parameterized by $k$) that provides a {\em recipe} for bounding the error probability of a $k$-ary sequential composition. A user only needs to prove the premise of the rule once (using constraint solving as described above) and can then apply the rule to obtain a bound on error probability for any desired $k$.
We are now ready to present our main {\em acceleration rule} that 
generalizes the preceding rules:

$$
  \linfer[Rule 3]{\{\phi\}C_1\{\phi\}\{\epsilon\}, \ldots, \{\phi\}C_m\{\phi\}\{\epsilon\}}
  {\{\phi\}(C_1 \barsp \ldots \barsp C_m)^k\{\phi\}\{1-(1-\epsilon)^k\}} 
$$
Using Rule 3, we can estimate the error probability for scenarios formed through arbitrary sequential compositions of $C_1$, $C_2$,  $C_m$ up to length $k$ 
 (here we use `$|$' to represent choice). 
 Like Rule 2, Rule 3 can be seen as a family of rules; it provides a {\em recipe} for bounding the error probability for sequences of scenarios of length $k$ that can be generated from the regular expression $(C_1 \barsp \ldots \barsp C_m)^k$.

 Note also that while the above rules are sound (see \Cref{app:soundness} for proofs) the error bound is not {\em tight}.
 Nevertheless, the acceleration rules give a domain expert the flexibility to efficiently bound the error probability for scenario sequences, without \textit{a priori} knowing the sequence length or the order of scenarios.

\subsubsection{Choice of $\phi$ and $\epsilon$.}\label{sec:finding-recurrences}
To apply Rule 3, we must find a precondition $\phi$ and $\epsilon \in \mathbb{R}$ that satisfy the rule's premise. This $\phi$ needs to be ``invariant'' in the sense that $\phi$ serves as both a pre- and post-condition. We can use Z3 to check whether a particular $\phi$ satisfies the rule's premise as described in the context of our example above. 
Before introducing our heuristic search to find an invariant $\phi$, we note that the vacuous precondition $\phi = \top$ satisfies Rule 3's premise $\{\phi\}C_1\{\phi\}\{\epsilon\}, \ldots, \{\phi\}C_m\{\phi\}\{\epsilon\}$ when $\epsilon$ is chosen as:\begin{equation*}
    \epsilon = \max_{i \in \countings{m}}\left(||\vb_i||_{\infty} \right)
\end{equation*}
where $C_i \equiv (\mA_i, \vb_i)$ for each $i$.
This $\epsilon$ is the worst-case \textit{local} error probability achieved from any initial distribution by any $C_i$.
In practice, this $\epsilon$ may be too high to provide a useful safety guarantee.

We describe our heuristic search for an invariant $\phi$ that allows us to apply Rule 3 to bound the error probability of arbitrary sequential compositions of a finite set of scenario sequences with summaries $(\mA_1, \vb_1), \ldots,(\mA_m, \vb_m)$. We use the same procedure in the context of Rule 2, which is equivalent to Rule 3 when $m=1$. Our search considers only the preconditions that are \textit{weakest preconditions} with respect to a particular value of $\epsilon$.
\begin{enumerate}
    \item Choose desired local error probability $0 \le \epsilon \le 1$ and define $\phi \coloneqq \land_{i \in \countings{m}}\vx \cdot \vb_i \le \epsilon$.
    \item Check whether $\forall i \in \countings{m}, \forall \vx \in \distr{\actualSys}, \phi(\vx) \Rightarrow \phi(\frac{\vx\mA_i}{1-\vx \cdot \vb_i})$ using a numerical solver such as Z3. 
    \item If the check from the previous step succeeds, then $\phi$ is an invariant precondition and we are done. Otherwise, we return to step 1 and choose a different, higher value of $\epsilon$.
\end{enumerate}

\section{Experimental Evaluation}\label{sec:experiments}
We apply our analysis techniques to two case studies, one based on the real-world TaxiNet perception system and another based on a simulated F1Tenth race car.
\subsubsection{TaxiNet}
TaxiNet is a neural network used for perception in an experimental system developed by Boeing for autonomous centerline tracking on airport taxiways. It takes as input a picture of taxiway and estimates the plane’s position with respect to the centerline in terms of two outputs: \textit{cross track error} (\textit{cte}), which is the distance in meters
of the plane from the centerline and\textit{ heading error} (\textit{he}), which is the angle 
of the plane with respect to the centerline. 
This regression is performed by a DNN containing 24 layers including five convolution layers and three dense layers (with 100/50/10 ELU neurons) before the output layer.
The resulting state estimates are fed to a controller, which maneuvers the plane to follow the centerline. Error is defined as excessive deviation from the centerline. We analyze a discretized version of the system, for details see \Cref{app:taxinet-details}. \

Prior work~\cite{pasareanu2023closed} builds a probabilistic abstraction of perception based on the performance of TaxiNet on a dataset provided by Boeing, and performs an analysis of the closed-loop system to estimate the probability of error
In our current work, we refine this analysis by first partitioning the dataset into a \textit{bright} dataset and a \textit{dark} dataset, then building a separate probabilistic abstraction of perception for each condition. 

\begin{table}[t]

\begin{minipage}[]{0.48\textwidth}
\centering
\textbf{TaxiNet Accuracy (Bright)} \\[0.5em]
\begin{tabular}{|c|c|c|c|c:c}

\cline{1-5}
 & \textbf{(0,0)} & \textbf{(0,1)} & \textbf{(0,2)} & \textbf{(1,0)} & \multirow{5}{*}{$\;\bm{\vdots}$} \\\cline{1-5}
\textbf{(0,0)} & 964 & 44 & 30 & 18 & \\\cline{1-5}
\textbf{(0,1)} & 4 & 233 & 0 & 2 & \\\cline{1-5}
\textbf{(0,2)} & 3 & 1 & 258 & 0 & \\\cline{1-5}
\textbf{(1,0)} & 45 & 39 & 3 & 475 & \\\cdashline{1-5}

\multicolumn{5}{c}{$\bm{\cdots}$} & \\
\end{tabular}

\end{minipage}
\hfill
\begin{minipage}[]{0.48\textwidth}
\centering
\textbf{TaxiNet Accuracy (Bright)} \\[0.5em]
\begin{tabular}{|c|c|c|c|c:c}
\cline{1-5}
 & \textbf{(0,0)} & \textbf{(0,1)} & \textbf{(0,2)} & \textbf{(1,0)} & \multirow{5}{*}{$\;\bm{\vdots}$} \\\cline{1-5}
\textbf{(0,0)} & 499 & 621 & 14 & 13 & \\\cline{1-5}
\textbf{(0,1)} & 1 & 473 & 0 & 3 & \\\cline{1-5}
\textbf{(0,2)} & 188 & 23 & 84 & 0 & \\\cline{1-5}
\textbf{(1,0)} & 203 & 765 & 1 & 125 & \\\cdashline{1-5}

\multicolumn{5}{c}{$\bm{\cdots}$} & \\
\end{tabular}

\end{minipage}

\caption{\small Excerpted rows and columns from the TaxiNet \textit{bright} (left) and \textit{dark} (right) confusion matrices. The rows and columns are labeled by states, expressed as (\textit{cte},\textit{he}) pairs.}
\vspace{-1cm}
 \label{tab:taxinet-cm}
\end{table}

\Cref{tab:taxinet-cm} shows excerpts from the confusion matrices made from the \textit{bright} and \textit{dark} partitions. Interestingly, the frequency and kind of misclassifications is markedly different between lighting conditions.
To investigate how different lighting conditions (and transitions between lighting conditions) affect system level safety, we construct two scenarios $\mathtt{bright} = (\mathit{bright}, 20)$ and $\mathtt{dark} = (\mathit{dark}, 20)$ that represent execution of the autonomous system for $20$ timesteps in \textit{bright} or \textit{dark} lighting, respectively and apply the compositional analysis techniques introduced in \Cref{sec:analysis}. To provide a point of comparison with prior work that builds a single probabilistic abstraction,~\cite{pasareanu2023closed}, we also consider a \textit{pooled} environment condition, which corresponds to the lighting condition being identically distributed at each timestep, proportionately to the frequency of light vs. dark images in the original, unpartitioned dataset. We define the scenario $\mathtt{pooled} = (\mathit{pooled}, 20)$ to demonstrate that our \texttt{bright} and \texttt{dark} scenarios permit a refined analysis.

\subsubsection{F1Tenth}\label{sec:results-f1tenth}
Our second case study is inspired by the F1Tenth autonomous racing competition~\cite{kelly2020f1tenth}, in which a model race car navigates a track. In particular, we analyze an idealized discrete-state model of driving dynamics.  The car's \textit{Sensor} produces 21-dimensional LiDAR observations, 
which are processed by a multilayer perceptron with ReLU activations, two hidden layers, and 128 neurons per hidden layer, to produce state estimates. 
We trained the MLP using $1000$ simulated observations gathered from each system state in each track segment and evaluated its performance using a separate, identically constructed dataset.
Each simulated LiDAR observation was constructed from a position sampled uniformly within a small L1 radius of the discrete state's canonical position.

Inspired by~\cite{ivanov2021compositional}, we consider the set of environment conditions to be the set of track segments, i.e. $\conditionSet = \{\mathit{left}, \mathit{right}, \mathit{straight}\}$ and define the scenarios $\mathtt{left} = (\mathit{left}, 30)$, $\mathtt{right} = (\mathit{right}, 30)$, and $\mathtt{straight} = (\mathit{straight}, 30)$.
We allow the \textit{Dynamics} $f : \conditionSet \times \actualSet \times \controlSet \rightarrow \actualSet$ and \textit{Controller}  $g : \conditionSet \times \estSet \rightarrow \controlSet$ to condition their behavior on the current environment condition. The dependency of $f$ on the current track segment is necessary to model the track configuration; the dependency of $g$ on the current track segment corresponds to a controller equipped with a perfect mode predictor that identifies the current scenario.

\begin{figure}[t]
    \centering
    \includegraphics[width=0.65\textwidth]{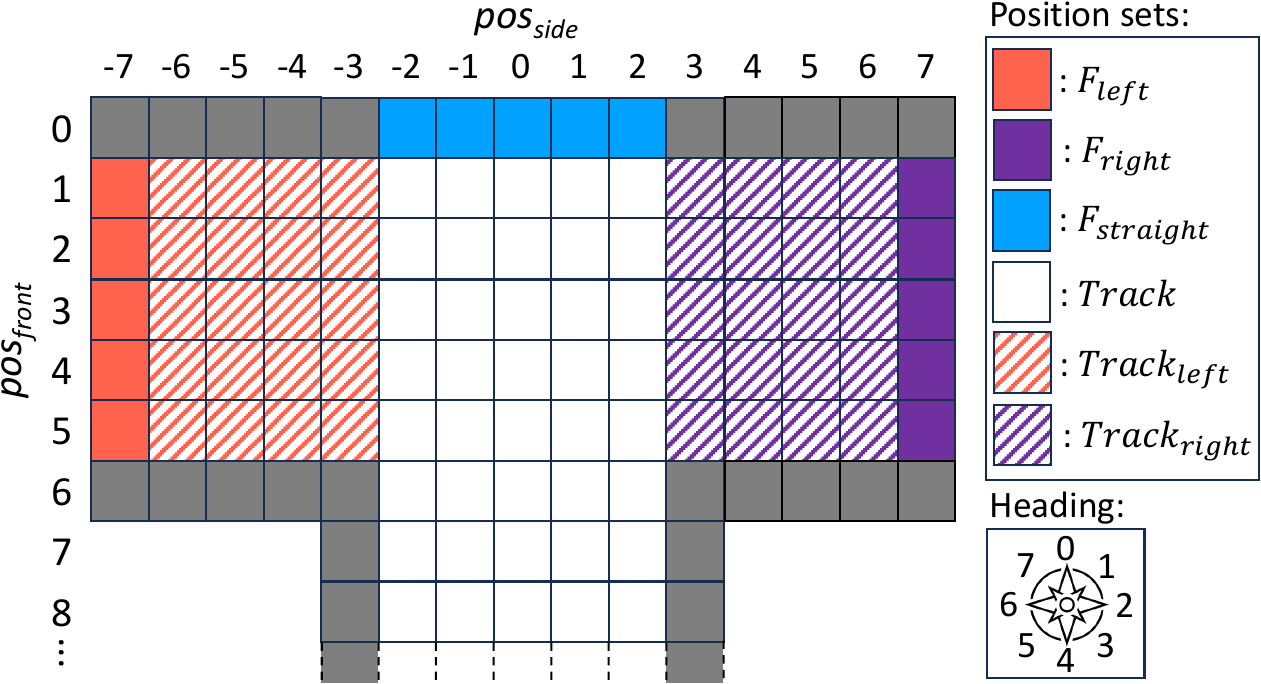}
    \caption{\small Each F1Tenth scenario corresponds to a track segment.}\vspace{-0.5cm}
    \label{fig:f1tenth}
\end{figure}

The meaning of a state $(\distS, \distF, \head, \tau) \in \actualSet \setminus \{\actualErr\}$ is defined with respect to the current scenario and diagrammed in \Cref{fig:f1tenth}. The components are:\begin{itemize}
    \item $\distS \in \{-7,\ldots,7\}$, which denotes sideways position.
    \item$\distF \in \{0,\ldots,16\}$, which denotes front/back position.
    \item $\head \in \{0, \ldots 7\}$ which denotes the car's heading, measured in increments of $\pi/4$ radians where $\head = 0$ denotes facing towards the front wall.
    \item $\tau \in \{1, \ldots, 30\}$ which denotes the timesteps spent in the current scenario.
\end{itemize}
The set of control inputs are $\controlSet = \{-1,0,1\}$ which represent adjustments to the current heading.
At each timestep, the car's heading is updated based on the control input $\control$. Then, the car moves one grid-position in the direction of the updated heading. 

To model walls and the final goal of each track segment, we define the subsets of the car positions as shown in \Cref{fig:f1tenth}. During the \texttt{left} scenario, the car's goal is to reach the final region $F_{\textit{left}}$. Failure to reach the final region $F_{\mathit{left}}$ in $30$ timesteps triggers a transition to $\actualErr$, as does attempting to enter any position not in $\mathit{Track} \cup \mathit{Track}_{\mathit{left}}$. This is defined similarly for the other track segments, for complete details, including technical treatment of the transitions between track segments, see~\Cref{app:f1-details}.

\subsection{Implementation}\label{sec:implementation}
We adapt the PRISM model construction of~\cite{pasareanu2023closed} to allow us to compute a \textit{summary} for each atomic scenario. The PRISM construction reflects the system decomposition illustrated in~\Cref{fig:components}. In particular, the PRISM model takes three timesteps (one for the learning-enabled components, one for the \textit{Controller}, and one for the \textit{Dynamics}) to model one timestep of the autonomous system. This coordination is mediated by bookkeeping variables, for details, see~\Cref{app:prism-model}.

Consider an autonomous system with state set $\actualSet$ and the atomic scenarios $(\condition_1, H_1), \ldots, (\condition_m, H_m)$. Our PRISM model contains a set of variables $V_\actualSet$ such that each valuation to $V_\actualSet$ uniquely determines a system state $\actual \in \actualSys$. Our model also contains a variable \texttt{scenario} that maintains the index $i \in \countings{m}$ of the current scenario and a variable \texttt{timer} that maintains the timestep of the original autonomous system.
To model the horizon of each scenario we define the PRISM formula \texttt{exit\_condition} as \texttt{! error \& ((scenario = 0 \& timer = H\_i) | \textit{...} | (scenario = m \& timer = H\_m))} and ensure that each PRISM state that satisfies \texttt{exit\_condition} has no outgoing transitions. To compute the summary $(\mA_i, \vb_i)$ of scenario $(\condition_i, H_i)$, we do the following for each $\actual \in \actualSys$:
\begin{enumerate}
        \item Modify the PRISM model so that initially $\mathtt{scenario}=i$, $\mathtt{timer}=1$ and the valuation to $V_\actualSet$ determines $\actual$.
    \item For each $\actual' \in \actualSet$, execute the PRISM query \texttt{P=? F[exit\_condition \& $\actual'$]} and fill this value into $\mA_i(\actual, \actual')$. In the preceding PRISM query, $\actual'$ stands for the state predicate that holds iff the valuation to $V_\actualSet$ determines $\actual'$.
    \item Execute the PRISM query \texttt{P=? F[error]} and fill this value into $\vb_i(\actual)$.
    \end{enumerate}

\subsection{Fixed Scenario Analysis}\label{sec:results-sequential}

We apply the analysis techniques described in \Cref{sec:analysis-sequential} to fixed sequences of the scenarios \texttt{bright} and \texttt{dark} from the TaxiNet case study to explore how a shift in lighting conditions affects system level safety. For the F1Tenth case study, we consider sequences of the scenarios \texttt{straight}, \texttt{left}, and \texttt{right} that represent different track configurations.

\def\taxiNominal{\phi_{\mathit{nominal}}^{\mathit{taxi}}}
\def\taxiCenter{\phi_{\mathit{center}}^\mathit{taxi}}
\def\taxiRight{\phi_{\mathit{right}}^\mathit{taxi}}

\def\ftenthPerfect{\phi_{\mathit{perfect}}^{\mathit{f1tenth}}}
\def\ftenthNominal{\phi_{\mathit{nominal}}^{\mathit{f1tenth}}}
\def\ftenthForward{\phi_{\mathit{forward}}^{\mathit{f1tenth}}}
\def\ftenthCenter{\phi_{\mathit{center}}^{\mathit{f1tenth}}}
\def\ftenthRight{\phi_{\mathit{right}}^\mathit{f1tenth}}

\begin{table}[t]
    \centering
    \begin{minipage}[]{0.49\textwidth}
        \centering
        \begin{tabular}{cccc}
            \toprule
            \multicolumn{4}{c}{\textbf{TaxiNet}} \\
            \midrule
            & $\taxiNominal$ & $\taxiCenter$ & $\taxiRight$ \\
            \midrule
            \texttt{bright;bright} & 0.030 & 0.099 & 0.215 \\ 
            \texttt{bright;dark} & 0.323 & 0.371 & 0.452 \\ 
            \texttt{dark;bright} & 0.368 & 0.570 & 0.455 \\ 
            \texttt{dark;dark} & 0.551 & 0.694 & 0.613 \\ 
            \texttt{pooled;pooled} & 0.307 & 0.462 & 0.399 \\ 
            \bottomrule
        \end{tabular}
        \centering
       
    \end{minipage}
    \hfill
    \begin{minipage}[]{0.49\textwidth}
        \centering
        \begin{tabular}{cccc}
            \toprule
            \multicolumn{4}{c}{\textbf{F1Tenth}} \\
            \midrule
            & $\ftenthNominal$ & $\ftenthCenter$ & $\ftenthRight$ \\
            \midrule
            \texttt{l;l;l;l}& 0.338 & 0.586 & 0.804\\ 
            \texttt{r;r;r;r} & 0.069 & 0.118 & 0.213 \\ 
            \texttt{s;s;s;s}& 0.00 & 0.00 & 0.00 \\ 
            \texttt{r;s;s;l}& 0.514 & 0.522 & 0.513\\ 
            \texttt{r;s;s;r} & 0.060  & 0.110 & 0.203 \\ 
            \bottomrule
        \end{tabular}
        \centering

    \end{minipage}
 
        \caption{\small Forward analysis for TaxiNet (left) and F1Tenth (right). For each scenario sequence, we report the worst-case error probability given a precondition over the initial state distribution. In F1Tenth, \texttt{l}, \texttt{r}, and \texttt{s} abbreviate \texttt{left}, \texttt{right}, and \texttt{straight}, respectively.}  \vspace{-1cm}
        \label{tab:sequential}
\end{table}

\subsubsection{Forward analysis} For the \textit{forward analysis}, we define preconditions over the initial state distribution and report the worst-case error probability in \Cref{tab:sequential}. For TaxiNet, we consider the precondition $\taxiNominal$ that requires the airplane to start with \textit{heading error} and \textit{cross track error} both equal to $0$ with probability $> 0.9$, the more lax precondition $\taxiCenter$ that merely requires the airplane to start close to the center of the runway with probability $> 0.9$, and the precondition $\taxiRight$ that requires the airplane to start on the far right side of the runway with probability $> 0.9$. We encode these preconditions as affine constraints.  For some boolean-valued expression $\psi$ over the variables $\mathit{cte}$ and $\mathit{he}$ we let $\vIndicator{\psi}$ denote the \textit{indicator vector} such that $\vIndicator{\psi}(\actual) = 1$ iff a valuation to PRISM variables that determines $\actual$ satisfies $\psi$.
\begin{align*}
    \taxiNominal =& \vx \vIndicator{\mathit{cte} \neq 0 \lor \mathit{he} \neq 0} \leq 0.1 \\
    \taxiCenter =& \vx\vIndicator{\mathit{cte} > 2} \leq 0.1\\
    \taxiRight  =& \vx\vIndicator{\mathit{cte} \neq 4} \leq 0.1
\end{align*}
We define the analogous preconditions $\ftenthNominal$, $\ftenthCenter$, and $\ftenthRight$ for the F1Tenth case study. As a technical detail, each F1Tenth precondition refines the precondition $\ftenthForward$, which ensures the car at the beginning of the track segment and facing within $\pi/4$ radians of straight forward with probability $1$. 
\begin{align*}
    \ftenthForward &= \vx  \vIndicator{|\distS|>2 \lor 1 <\head < 7 \lor \tau \neq 0 \lor \distF \neq 15} \leq 0 \\
    \ftenthNominal &= \ftenthForward \land \vx\vIndicator{\distS \neq 0 \lor \head \neq 0} \leq 0.1, \\
    \ftenthCenter &= \ftenthForward \land \vx\vIndicator{|\distS|>1} \leq 0.1, \\
    \ftenthRight  &= \ftenthForward \land \vx\vIndicator{\distS \neq 2} \leq 0.1.
\end{align*}

\paragraph{Environment conditions.} In the TaxiNet case study, the sequence \texttt{bright;bright} has low error probability, which shows that the learning-enabled component and controller work well in well-lit conditions. On the other hand, each sequence that includes the \texttt{dark} scenario has extremely high error probability. 
This dependence on light conditions would not have been detectable if we had not built separate probabilistic abstractions of perception for light and dark operating conditions. We include hypothetical results based on the naive probabilistic abstraction of perception that pools data collected in both bright and dark conditions as the scenario sequence $\mathtt{pooled};\mathtt{pooled}$. 

Similarly, we observe that when started from an initial state distribution that satisfies $\ftenthNominal$, the F1Tenth car achieves low worst case error probability (0.069) for the track configuration \texttt{r;r;r;r} but a relatively high worst case error probability (0.338) for \texttt{l;l;l;l}. A system designer could use this information to understand that it may be acceptable to deploy the car on a track that loops clockwise, but the system should be retooled before deployment on a counterclockwise loop.

\paragraph{Initial distributions.} Our forward analysis also shows how the initial state distribution affects system-level safety. This effect is best illustrated in the F1Tenth case study. 
Returning to \Cref{tab:sequential}, we observe that the precondition $\ftenthNominal$ ensures the lowest error probability of any of the considered preconditions for each scenario that we consider. Some scenarios, namely namely \texttt{s;s;s;s} and \texttt{r;s;s;l} are relatively robust to less-than-ideal initial distributions and exhibit a similar worst-case error probability for the more lax precondition $\ftenthCenter$ as they do for the strict $\ftenthNominal$.  Interestingly, we also observe that $\ftenthRight$, which requires the car to start on the right side of the track with probability $> 0.9$ leads to extremely high error probability for the scenario \texttt{l;l;l;l} in which the car must navigate a counterclockwise loop. Our forward analysis can help a system designer discover which combinations of track segments and initial state distributions will permit safe deployment of the autonomous system. The results from our F1Tenth case study would cause a system designer to focus on improving performance on track configurations that include \texttt{left} turn segments.

\subsubsection{Backward analysis} 
\begin{figure}[t]
    \begin{center}
   \includegraphics[width=\textwidth]{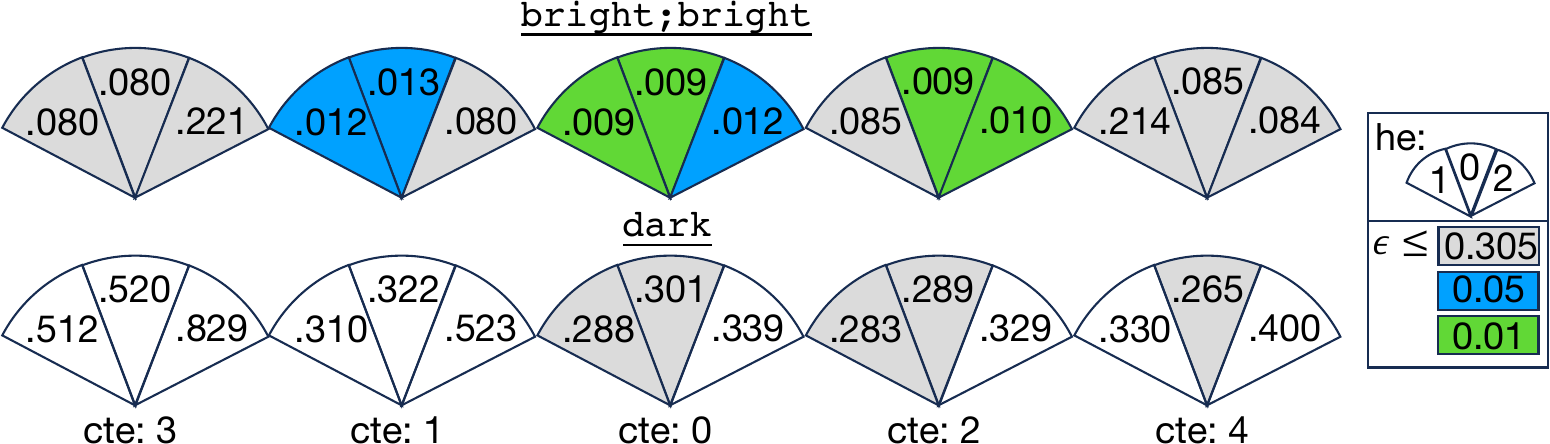}
    \end{center}\vspace{-0.5cm}
    \caption{\small Error probability by initial point distribution for \texttt{bright;bright} and \texttt{dark} scenarios. Initial point distributions that ensure error probability at most $0.305$, $0.05$ and $0.01$ are highlighted in gray, blue, and green, respectively. }\vspace{-0.5cm}
    \label{fig:taxinet-sequential}
\end{figure}

For the \textit{backward analysis}, we assume that the domain expert provides a scenario sequence along with a maximum allowable error probability $\epsilon$ for the entire execution of the scenario sequence. We then compute the summary $(\mA,\vb)$ of the scenario sequence and observe that the weakest precondition that ensures the error probability does not exceed $\epsilon$ is $\phi_\epsilon \coloneqq \vx\cdot\vb\leq\epsilon$. 
We visualize these preconditions in 
\Cref{fig:taxinet-sequential} by highlighting the set of point distributions that ensure error probabilities less than $0.305$, $0.05$, and $0.01$ for the scenario sequence \texttt{bright;bright} and the (singleton) scenario sequence \texttt{dark}.

\subsubsection{Computational overhead}\label{sec:results-parametric-mc}
Both the \textit{forward} and \textit{backward} analyses are very computationally efficient, assuming that the summary of each atomic scenario has already been computed.  The linear programs generated during our forward analysis can be quickly solved by Gurobi; it took less than $0.2$ seconds to compute all the values in \Cref{tab:sequential}. The backward analysis requires no computation beyond the construction of $\vb$. The upfront summary computation only needs to be performed once per atomic scenario. Generating all summaries for the TaxiNet and F1Tenth case studies took 79.56 seconds and 360.86 seconds, respectively.\footnote{All durations are reported based on single-threaded execution using a commodity laptop with a 2 GHZ processor.} 

\begin{figure}[t]
    \begin{center}
   \includegraphics[width=\textwidth]{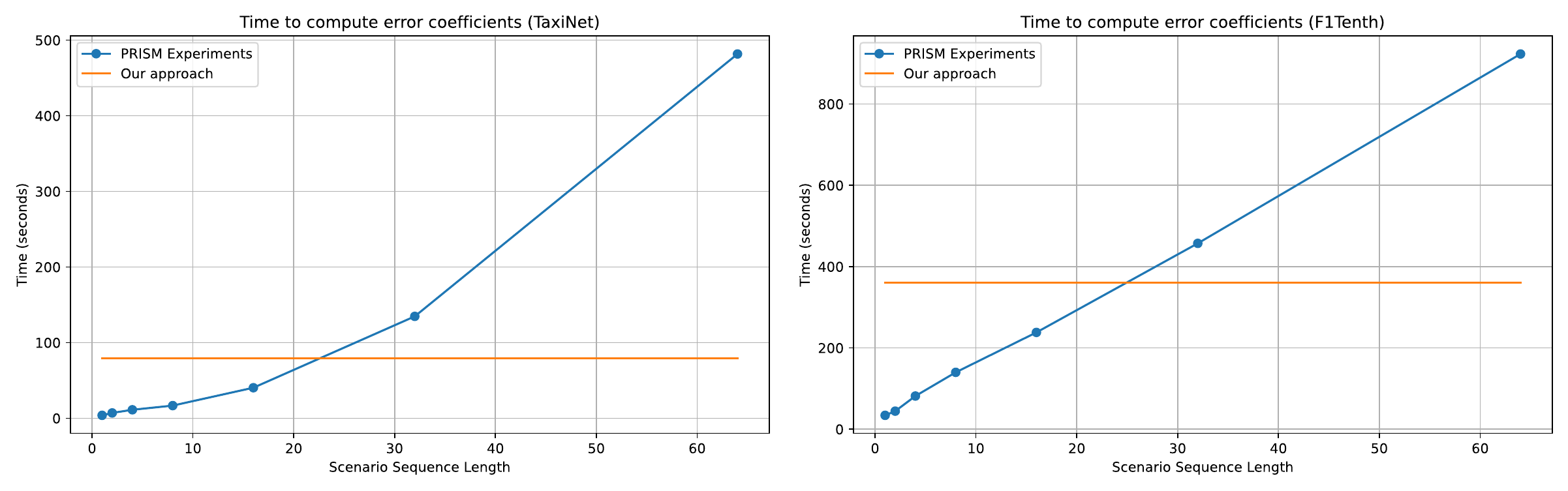}
    \end{center}\vspace{-0.5cm}
    \caption{\small Time needed to construct the error coefficients vector $\vb$ using our compositional approach vs. a non-compositional application of PRISM's Experiments feature. For our approach, we report the time needed for the one-time summary generation, which can then be composed with minimal overhead for any sequence of scenarios. For PRISM experiments, we report the time needed to compute $\vb$ for fixed scenario sequences of varying length. For TaxiNet we chose scenario sequences containing only \texttt{bright}, for F1Tenth we chose scenario sequences containing only \texttt{right}.}\vspace{-0.5cm}
    \label{fig:computational-overhead}
\end{figure}

Our analyses consider \textit{sets} of initial distributions, which are not naturally expressible in PRISM, so a direct comparison of computational efficiency vs. PRISM is not possible. However, one could use PRISM compute the error probability coefficients $\vb$ by first building a monolithic DTMC that represents the execution of a particular scenario, then using either (1) PRISM's parametric model checking feature or (2) PRISM's experiments feature. Neither method scales gracefully: for our TaxiNet case study using parametric model checking to compute $\vb$ for the scenario \texttt{bright} takes 6.912 seconds and exceeds PRISM's default 1GB of allocated RAM for \texttt{bright; bright}. PRISM's experiments feature can calculate $\vb$ for \texttt{bright;bright} in 7.027s, however computing $\vb$ for the long scenario \hbox{$\mathtt{bright}^{64}$} takes 481.692s much more than the $\sim 79.56$ seconds needed by our approach. We visualize the computational scalability in \Cref{fig:computational-overhead}.

\subsection{Using the Acceleration Rule}\label{sec:results-compositional}

The acceleration rules introduced in \Cref{sec:analysis-compositional} allow us to bound error probability of sequence scenarios, without putting an \textit{a priori} bound on sequence length.
Empirically, we find that our procedure to guess invariant preconditions works well for our case studies.
Out of all of our singleton scenario sequences (\texttt{bright}, \texttt{dark}, \texttt{straight}, \texttt{left}, and \textbf{right}) and the $\epsilon$ values $\{0, 0.01, 0.02, \ldots 0.99 \}$, the only instances where our guessed $\phi_\epsilon$ did not satisfy the premise of Rule 2 were:\begin{itemize}
    \item TaxiNet's \texttt{dark} scenario with the $\epsilon$ values $0.27$, $0.28$, $0.29$, and $0.30$.
    \item F1Tenths's \texttt{left} scenario with $0.02 \leq \epsilon \leq 0.13$.
    \item F1Tenth's \texttt{right} scenario with $\epsilon$ values $0.00$, $0.01$, and $0.02$.
\end{itemize}

\subsubsection{Acceleration.} Practically, the procedure yields an upper bound on the error probability of an unbounded sequential iteration of a single scenario. For example, for a desired \textit{local} error probability of $\epsilon = 0.01$, we apply Rule 2 to TaxiNet's \texttt{bright} scenario with $\phi = \vx \cdot  \vb_{\mathtt{bright}}\leq 0.01 $ to prove:
$$
  {\{\phi\}(\mathtt{bright})^k\{\phi\}\{1-(1-0.01)^k\}} 
$$
for any $k \in \mathbb{N}$. This bound on error probability is parametric in the number of iterated executions of the \texttt{bright} scenario. Though not necessarily tight, such a parametric upper bound can help a system designer reason about how the cumulative probability of error increases during the execution of the autonomous system.

Rule 2 is particularly useful for autonomous systems that can maintain perfect safety. In our F1Tenth case study, we found that for the \texttt{straight} scenario, the precondition $\phi = \vx \cdot \vb_{\texttt{straight}} \le 0$ is invariant, so we can prove:
$$
  {\{\phi\}(\mathtt{straight})^k\{\phi\}\{0\}} 
$$
for any $k$. The precondition $\phi$ is actually quite permissive, in particular, it is more permissive than any of $\ftenthNominal$, $\ftenthCenter$, and $\ftenthRight$ introduced in $\Cref{sec:analysis-sequential}$.

\subsubsection{Acceleration with choice.} 
\begin{figure}[t]
    \begin{center}
   \includegraphics[width=\textwidth]{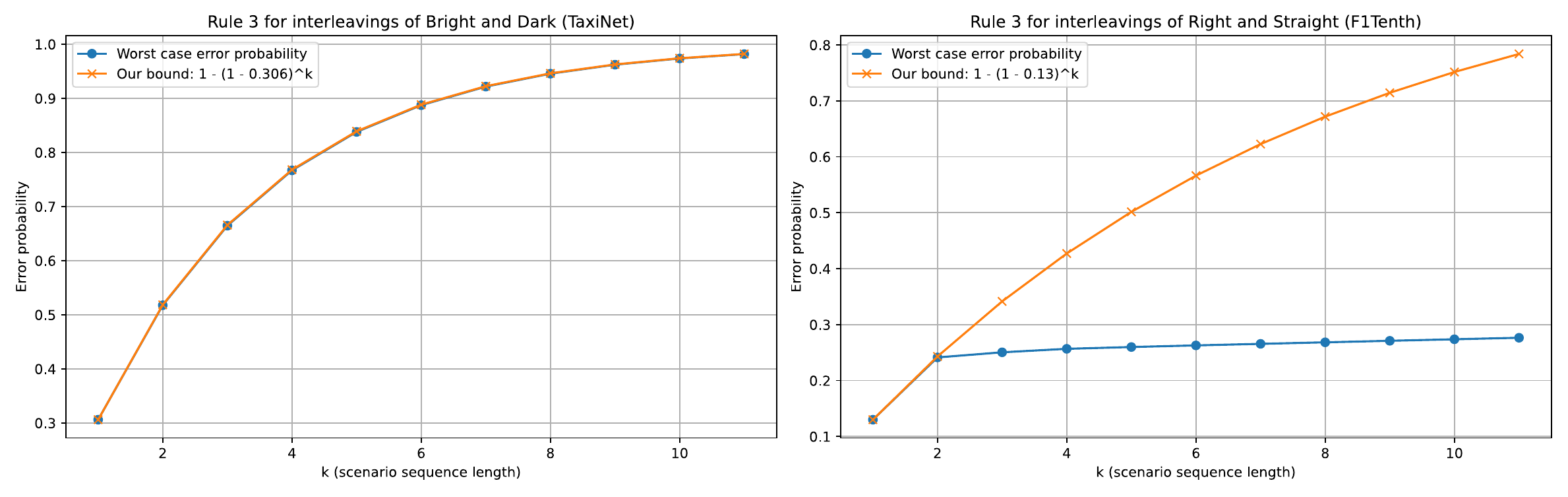}
    \end{center}\vspace{-0.5cm}
    \caption{\small The error bounds found by applications of Rule 3 vs. true worst-case error probability. For the worst case error probabilities, we consider all possible scensario interleavings and all starting distributions that satisfy the precondition $\phi_{\epsilon}$ used in our application of Rule 3.}\vspace{-0.5cm}
    \label{fig:rule3}. 
\end{figure}

For the more complex Rule 3, once again, we found that our proposed procedure successfully finds recurrent preconditions in the context of our case studies. For the TaxiNet case study, applying this procedure for $\epsilon = 0.306$ and $\phi_{0.306} \coloneqq  \vx \cdot \vb_{\mathtt{bright}} \le 0.306 \land  \vx \cdot \vb_{\mathtt{dark}} \le 0.306$ allows us to prove: 
$$
  {\{\phi_{0.306}\}(\mathtt{bright} \barsp \mathtt{dark})^k\{\phi_{0.306}\}\{1-(1-0.306)^k\}} 
$$
this form of error bound is useful to a system designer who cannot predict the sequence of atomic scenarios the autonomous system will encounter during operation.
We plot this upper bound on error probability against the true error probability of an adversarially chosen scenario sequence in \Cref{fig:rule3}.

In the context of the F1Tenth case study, Rule 3 lets us derive an error bound for arbitrary track configurations. Importantly, we only need to collect datasets for each atomic scenario and our compositional reasoning allows us to generalize our guarantee to any track configuration formed from these atomic scenarios. Trying our procedure to guess an invariant precondition that satisfies the premise of Rule 3 for the scenarios \texttt{left}, \texttt{right}, and \texttt{straight}, we found that for $\epsilon \in \{0.74, 0.75, \ldots, 0.99\}$ the precondition $\phi_\epsilon \coloneqq \vx \cdot \vb_\mathtt{left} \le \epsilon \land \vx \cdot \vb_\mathtt{right} \le \epsilon \land \vx \cdot \vb_\mathtt{straight} \le \epsilon$ is invariant and nontrivial. We can thus conclude:
$$
    {\{\phi_{0.74}\}(\mathtt{left} \barsp \mathtt{right} \barsp \mathtt{straight})^k\{\phi_{0.74}\}\{1-(1-0.74)^k\}}
$$
This bound allows extremely high error probability. A domain expert might restrict the deployment of the autonomous car to tracks formed from \texttt{right} and \texttt{straight} segments. Here, we find that $\phi_\epsilon \coloneqq \vx \cdot \vb_\mathtt{right} \le \epsilon \land \vx \cdot \vb_\mathtt{straight} \le \epsilon$ is nontrivial and invariant for $\epsilon \in \{0.13, 0.14, \ldots, 0.99\}$. In particular, we can conclude 
$$\{\phi_{0.13}\}(\mathtt{right} \barsp \mathtt{straight})^k \{\phi_{0.13}\}\{1-(1-0.13)^k\}$$
Concerning the relatively high error probabilities, we currently work with high-level abstractions of actual systems so computed error probabilities should be taken with a grain of salt. Still, our acceleration rules can reveal useful trends to system developers. Moreover, we expect lower error probabilities when applied to more accurate models of real-world systems; the analysis of fixed scenario sequences (\Cref{sec:analysis-sequential}) is tight.

\section{Conclusion}

Our verification framework 
decomposes the autonomous system 
and its operating conditions into scenarios to enable efficient probabilistic analysis. 
Our compositional proof rules enable system designers 
to obtain a bound on error probability that is parametric with respect to the length of an arbitrary interleaving of scenarios, which is useful for reasoning about the system under unpredictable changes in operating conditions.
In future work, we plan to investigate more nuanced approaches to discover the invariant preconditions that enable application of our acceleration rule. 

Building a probabilistic abstraction of perception with respect to a discretized state space can introduce inaccuracies. In our work, we avoid this concern by assuming that the distribution of state estimates yielded by the learning-enabled component is uniquely determined by the current environment condition and (discrete) state. In future work we plan to develop (possibly symbolic) abstractions of perception for continuous-state system models and to account for distribution shifts that are continuous, instead of discrete as we do here.

\bibliographystyle{splncs04}
\bibliography{main,surveybib}

\appendix

\section{Soundness Proofs}\label{app:soundness}
We prove soundness of the rules introduced in \Cref{sec:analysis-compositional} with respect to the definition of sequential composition of \textit{summaries} found at the end of \Cref{sec:scenario}.

$$
  \linfer[Rule 1]{\{\phi\}C\{\psi\}\{\epsilon\}, \{\phi'\}C'\{\psi'\}\{\epsilon'\}, \psi\implies \phi'}
  {\{\phi\}C C'\{\psi'\}\{1-(1-\epsilon)(1-\epsilon')\}} 
$$
\begin{proof}
Let $C\equiv (\mA,\vb)$ and $C'\equiv (\mA',\vb')$.\\
By the definition of sequential composition of scenarios, we know $CC' \equiv (\mA\mA', \vb+\mA\vb$.\\
We prove the two obligations:\begin{itemize}
    \item To show that $\phi(\vx)\Rightarrow \psi'(\norm{\vx\mA\mA'})$ we fix $\vx$ w.l.o.g. and apply the first premise to obtain $\psi(\mathit{norm}(\vx\mA))$. We then apply the third premise to obtain $\phi(\mathit{norm}(\vx\mA))$ and the second premise to obtain $\psi'(\mathit{norm}(\mathit{norm}(\vx\mA)\mA'))$. Expansion of the definition of \textit{norm} and algebraic manipulation completes this branch of the proof.
    \item To show that $\phi(\vx) \rightarrow \vx \vb +\vx\mA\vb' \le 1-(1-\epsilon)(1-\epsilon')$ we fix $\vx$ w.l.o.g.
    Next we observe that $\vx \vb +\vx\mA\vb' = \vx \vb +(1-\vx\vb)\vx\mA\vb'$ by the definition of norm and of a summary. We can rewrite this as $1-(1-\vx\vb)(1-\norm{\vx\mA}\vb')$. So it suffices to show $1-(1-\vx\vb)(1-\norm{\vx\mA}\vb') \le 1-(1-\epsilon)(1-\epsilon')$. Premise 1 lets us bound $\vx\vb \le \epsilon$ and premise 2 lets us bound $\norm{\vx\mA}\vb' \le \epsilon'$ so we can bound $(1-\vx\vb) \ge 1-\epsilon$ and $1-\norm{\vx\mA}\vb' \ge 1-\epsilon'$. Together these imply the desired bound.

\end{itemize}
\end{proof}

$$
  \linfer[Rule 2]{\{\phi\}C\{\phi\}\{\epsilon\}}
  {\{\phi\}C^k\{\phi\}\{1-(1-\epsilon)^k\}} 
$$
\begin{proof}
    Direct application of Rule 3, which we prove below.
\end{proof}
$$
  \linfer[Rule 3]{\{\phi\}C_1\{\phi\}\{\epsilon\}, \ldots, \{\phi\}C_\ell\{\phi\}\{\epsilon\}}
  {\{\phi\}(C_1 \barsp \ldots \barsp C_m)^k\{\phi\}\{\epsilon\}\}} 
$$
\begin{proof}
    By induction over $k$. The case $k = 1$ is immediate and the case $k=2$ follows by Rule 1. For the inductive step, assume w.l.o.g. that the conclusion of this instance of Rule 3 is
    $\{\phi\}C_{i_1}\ldots C_{i_{k-1}}C_{i_k}\{\phi\}\{1-(1-\epsilon)^k\}$. By the inductive hypothesis, we know $\{\phi\}C_{i_1}\ldots C_{i_{k-1}}\{\phi\}\{1-(1-\epsilon)^{(k-1)}\}$, so we can apply Rule 1 to complete the proof.
\end{proof}

\section{PRISM Model Construction}\label{app:prism-model}
We provide additional details about the our PRISM model construction introduced in~\Cref{sec:implementation} and adapted from~\cite{pasareanu2023closed}.
Assume that we are modeling an autonomous system in which the dynamics uses state set $\actualSet$, the perception model outputs state estimates from $\estSet$, the controller issues commands from $\controlSet$, and there are $m$ distinct scenarios which we will refer to by the counting numbers $\countings{m}$. For each scenario $i$, we denote the probabilistic abstraction of perception as $\alpha_i : \actualSet \rightarrow \distr{\estSet}$ the time horizon as $H_i \in \mathbb{N}$. Let $H_{max}$ denote $\max_{i \in \countings{m}}(H_i)$.

Our PRISM model has a set of state variables $V$ which can be partitioned as $V = V_\actualSet \sqcup V_\estSet \sqcup V_\controlSet \sqcup \{\mathtt{pc}, \mathtt{scenario}, \mathtt{timer}\}$. Each valuation $V_\actualSet$, $V_\estSet$, or $V_\controlSet$ uniquely determines an element of $\actualSet$, $\estSet$, or $\controlSet$, respectively.
For each $\actual \in \actualSet$, we write $\actual$ as shorthand for the PRISM state predicate that evaluates to true exactly when the current valuation of $V_\actualSet$ determines $\actual$.

Our construction introduces a \textit{program counter} variable \texttt{pc} with range $\{0,1,2\}$, a variable \texttt{scenario} with range $\countings{m}$, and a variable \texttt{timer} with range $\countings{H_{\mathit{max}}}$. PRISM indexes timesteps starting at 0. We maintain \texttt{pc} and \texttt{timer} such that at PRISM timestep $t$, we have $\mathtt{pc} = \mathit{mod}(t,3)$ and $\mathtt{timer} = \min(\mathit{quotient}(t,3)+1, H_{\mathit{max}})$. The other variables are updated as follows:\begin{itemize}
    \item \textbf{Perception:} When $pc=0$, the valuation of $V_\estSet$ will be updated according to the valuation of $V_\actualSet$ and on the next step. This update's transition probabilities are given by the \textit{probabilistic abstraction} for the current scenario, which we represent using the value of \texttt{scenario}. The value of \texttt{scenario} is never updated, later on we will explain how we initialize the value of scenario to compute each summary.
    \item \textbf{Controller:} When $pc=1$, the valuation of $V_\controlSet$ is updated according to the valuation of $V_\estSet$.
    \item \textbf{Dynamics:} When $pc=2$, the valuation of $V_\actualSet$ is updated according to the valuation of $V_\controlSet$ and the current valuation of $V_\actualSet$.
\end{itemize}
We can now apply the summary generation technique detailed in~\Cref{sec:implementation} to this model.

\subsection{Additional TaxiNet Details}\label{app:taxinet-details}
For the purpose of our analysis, we discretize TaxiNet's outputs and treat it as a classifier.
cte $\in$ [-8.0 m, 8.0 m] and %
he $\in$ [-35.0 deg, 35.0 deg] %
are translated into cte $\in$ {0, 1, 2, 3, 4} and he $\in$ {0, 1, 2} as
shown below.

\noindent\begin{minipage}{.5\linewidth}
\begin{equation*}
\resizebox{0.9\hsize}{!}{
  $cte =
  \setlength{\arraycolsep}{0pt}
  \renewcommand{\arraystretch}{1.2}
  \left\{\begin{array}{l @{~} l} 
        3            & \text{if}~-8.0~\text{m}~ 
        <= cte < -4.8~\text{m} \\
        1            & \text{if}~-4.8~\text{m}~ 
        <= cte < -1.6~\text{m}  \\
        0            & \text{if}~-1.6~\text{m}~ 
        <= cte <= 1.6~\text{m}  \\
        2            & \text{if}~1.6~\text{m}~ 
        < cte <= 4.8~\text{m}  \\
        4            & \text{if}~4.8~\text{m}~ 
        < cte <= 8.0~\text{m} 
  \end{array}\right.$}
\end{equation*}
\end{minipage}%
\begin{minipage}{.5\linewidth}
\begin{equation*}
\resizebox{0.9\hsize}{!}{
  $he =
  \setlength{\arraycolsep}{0pt}
  \renewcommand{\arraystretch}{1.2}
  \left\{\begin{array}{l @{~} l} 
        1           & \text{if}~-35.0~ deg~ 
        <= he < -11.67~deg \\
        0           & \text{if}~-11.67 ~deg~ 
        <= he <= 11.66~deg \\
        2           & \text{if}~11.66 ~deg~ 
        < he <= 35.0~deg
  \end{array}\right.$}
\end{equation*}
\end{minipage}\vspace{0.3cm}
Yielding the discretized set of system states $\actualSys \coloneqq [0..4] \times [0..2]$. Any airplane position with in which the magnitude of the cross-track error exceeds 8m or the magnitude of the heading error exceeds $35^\circ$ represents the error state $\actualErr$.

\subsection{Additional F1Tenth Details}\label{app:f1-details}
We provide additional details about the F1Tenth case study.
At each timestep, the car's heading $\head$ is updated based on the control input $u$. Then, the car moves one grid-position in the direction of the updated heading. The dynamics function $f : \conditionSet \times \actualSet \times \controlSet$ also accounts for walls and transitions between track segments.

We define one scenario $(\condition, \maxt)$ for each track segment $\condition$. We wish for the scenario $(\condition, \maxt)$ to end when the car reaches the end of the current track segment, which we represent as $F_\condition$. This requires careful definition of the dynamics function, since the car may take more or less than the horizon value of $\maxt$ timesteps to reach $F_\condition$. We present pseudocode for the dynamics function in $f$. At a high level, we introduce a time limit $\maxt$ and transition to $\actualErr$ if the car does not reach $F_\condition$ in fewer than $\maxt$ steps. To address cars that reach $F_\condition$ in less than $\maxt$ timesteps, we stop updating the position and heading of the car until the start of the next scenario.

\SetKwInput{KwInput}{Input} 
\SetKwInput{KwOutput}{Output} 
\SetKwRepeat{Do}{do}{while}
\SetKwIF{If}{ElseIf}{Else}{if}{then}{else if}{else}{}
\begin{algorithm}[t]
\SetAlgoLined
\captionsetup{format=plain,labelfont=bf,justification=centering}
\caption{F1Tenth Dynamics $f: \conditionSet \times \actualSet \times \controlSet \rightarrow \actualSet$}
\label{alg:f1tenth}
\KwInput{Track segment $e \in E$; Current state $\actual \in \actualSet$; control input $\control \in \controlSet$}
\KwOutput{Next state $\actual' \in \actualSet$}

\uIf{$\actual = \actualErr$}{
    \Return{$\actualErr$}
}\Else{
    $(\distS,\distF,\head,\tau) \leftarrow \actual$\;
    \uIf{$\tau < 30$}{
        $\head' \leftarrow \mathit{mod}(\head + u, 8)$\;
        $\distS' \leftarrow \distS - \indicator_{\{1,2,3\}}(\head) + \indicator_{\{5,6,7\}}(\head)$\;
        $\distF' \leftarrow \distF - \indicator_{\{0,1,7\}}(\head) + \indicator_{\{3,4,5\}}(\head)$\;
        $(\distS', \distF') \leftarrow \fpos((\distS,\distF,\head),u)$\;
        \uElseIf{$(\distS',\distF') \in \mathit{Track} \cup \mathit{Track}_\condition \cup F_\condition$}{
            \Return{$(\distS',\distF',\head',\tau+1)$}\;
        }\Else{
            \Return{$\actualErr$}\;
        }
    }\uElseIf{$(\distS,\distF) \in F_\condition$}{
        \uIf{$\condition = \mathtt{left}$}{
            \Return{$(3-\distF,15,\mathit{mod}(\head - 2, 8),1)$}\;
        }\uElseIf{$\condition = \mathtt{right}$}{
            \Return{$(\distF-3,15,\mathit{mod}(\head + 2, 8),1)$}\;      
        }\Else{
            \Return{$(\distS,15,\head,1)$}\;
        }
    }\Else{
        \Return{$\actualErr$}
    }
}
\caption{F1Tenth Dynamics}
\end{algorithm}

We built a \textit{Controller} $g: \conditionSet \times \estSet \rightarrow \controlSet$. The dependence on $\estSet$ corresponds to the autonomous system being equipped with an oracle that detects which track segment is being navigated currently. 
This controller can navigate each track segment safely when it recieves the ground-truth state estimate at each timestep and begins from an initial state with heading $\head = 0$. When operating under these conditions, the controller completes each track segment with its heading such that it starts the next segment with $\head =0$. This implies that the controller can safely navigate any sequence of track segments assuming perfect behavior of the learning-enabled component.

\end{document}